\documentclass[10pt,twocolumn,letterpaper]{article}

\usepackage[pagenumbers]{cvpr} 

\usepackage{graphicx}
\usepackage{amsmath}
\usepackage{amssymb}
\usepackage{booktabs}

\usepackage{caption} 
\usepackage[font=small]{caption}

\usepackage{multirow}
\usepackage{xcolor}  
\usepackage{multirow}
\usepackage{siunitx}  
\usepackage{float}
\usepackage{arydshln}  
\usepackage{pifont}  
\usepackage{dblfloatfix} 

\usepackage[pagebackref,breaklinks,colorlinks]{hyperref}

\usepackage[capitalize]{cleveref}
\crefname{section}{Sec.}{Secs.}
\Crefname{section}{Section}{Sections}
\Crefname{table}{Table}{Tables}
\crefname{table}{Tab.}{Tabs.}

\newcommand{\datasetName}{MP-DocVQA\xspace}
\newcommand{\methodFullName}{Hierarchical Visual T5\xspace}
\newcommand{\methodName}{Hi-VT5\xspace}
\newcommand{\datasetPages}{$47,952$\xspace}
\newcommand{\clsToken}{{\fontfamily{pcr}\selectfont [CLS]}\xspace}
\newcommand{\pageToken}{{\fontfamily{pcr}\selectfont [PAGE]}\xspace}
\newcommand{\pageTokenC}{{\fontfamily{pcr}\selectfont [PAGE]\textsuperscript{'}}\xspace}

\newcommand{\longformer}{Longformer\xspace}
\newcommand{\bigbird}{Big Bird\xspace}

\newcommand{\setup}{setup\xspace}
\newcommand{\setups}{setups\xspace}

\newcommand{\oracle}{\textit{`oracle'}\xspace}
\newcommand{\concat}{\textit{`concat'}\xspace}
\newcommand{\logits}{\textit{`max conf.'}\xspace}
\newcommand{\logitsx}{max conf\xspace}

\newcommand{\MaxConf}{Max Conf.\xspace}

\newcommand{\multipage}{multi-page\xspace}
\newcommand{\multimodal}{multimodal\xspace}

\def\cvprPaperID{11503} 
\def\confName{CVPR}
\def\confYear{2023}

\begin{document}

\title{Hierarchical multimodal transformers for Multi-Page DocVQA}


\author{Rub{\`e}n Tito \and Dimosthenis Karatzas \and Ernest Valveny \\ \\
}

\affiliation{
Computer Vision Center, UAB \\
{\tt\small \{rperez, dimos, ernest\}@cvc.uab.es}
}


\maketitle


\begin{abstract}
   Document Visual Question Answering (DocVQA) refers to the task of answering questions from document images. Existing work on DocVQA only considers single-page documents. However, in real scenarios documents are mostly composed of multiple pages that should be processed altogether. In this work we extend DocVQA to the \multipage scenario. For that, we first create a new dataset, \datasetName, where questions are posed over \multipage documents instead of single pages. Second, we propose a new hierarchical method, \methodName, based on the T5 architecture, that overcomes the limitations of current methods to process long \multipage documents. The proposed method is based on a hierarchical transformer architecture where the encoder summarizes the most relevant information of every page and then, the decoder takes this summarized information to generate the final answer. 
   Through extensive experimentation, we demonstrate that our method is able, in a single stage, to answer the questions and provide the page that contains the relevant information to find the answer, which can be used as a kind of explainability measure.
\end{abstract}

\vspace{-8pt}
\section{Introduction} \label{sec:intro}
\vspace{-2pt}
Automatically managing document workflows is paramount in various sectors including Banking, Insurance, Public Administration, and the running of virtually every business. For example, only in the UK more than 1 million home insurance claims are processed every year.
Document Image Analysis and Recognition (DIAR) is at the meeting point between computer vision and NLP. For the past 50 years, DIAR methods have focused on specific information extraction and conversion tasks. 
Recently, the concept of Visual Question Answering was introduced in DIAR ~\cite{mathew2020document,mathew2021docvqa,mathew2022infographicvqa}. This resulted in a paradigm shift, giving rise to end-to-end methods that condition the information extraction pipeline on the natural-language defined task.
DocVQA is a complex task that requires reasoning over typed or handwritten text, layout, graphical elements such as diagrams and figures, tabular structures, signatures and the semantics that these convey.


\begin{figure}[ht]
    \centering
    \begin{subfigure}[b]{\linewidth}
        \includegraphics[width=\textwidth]{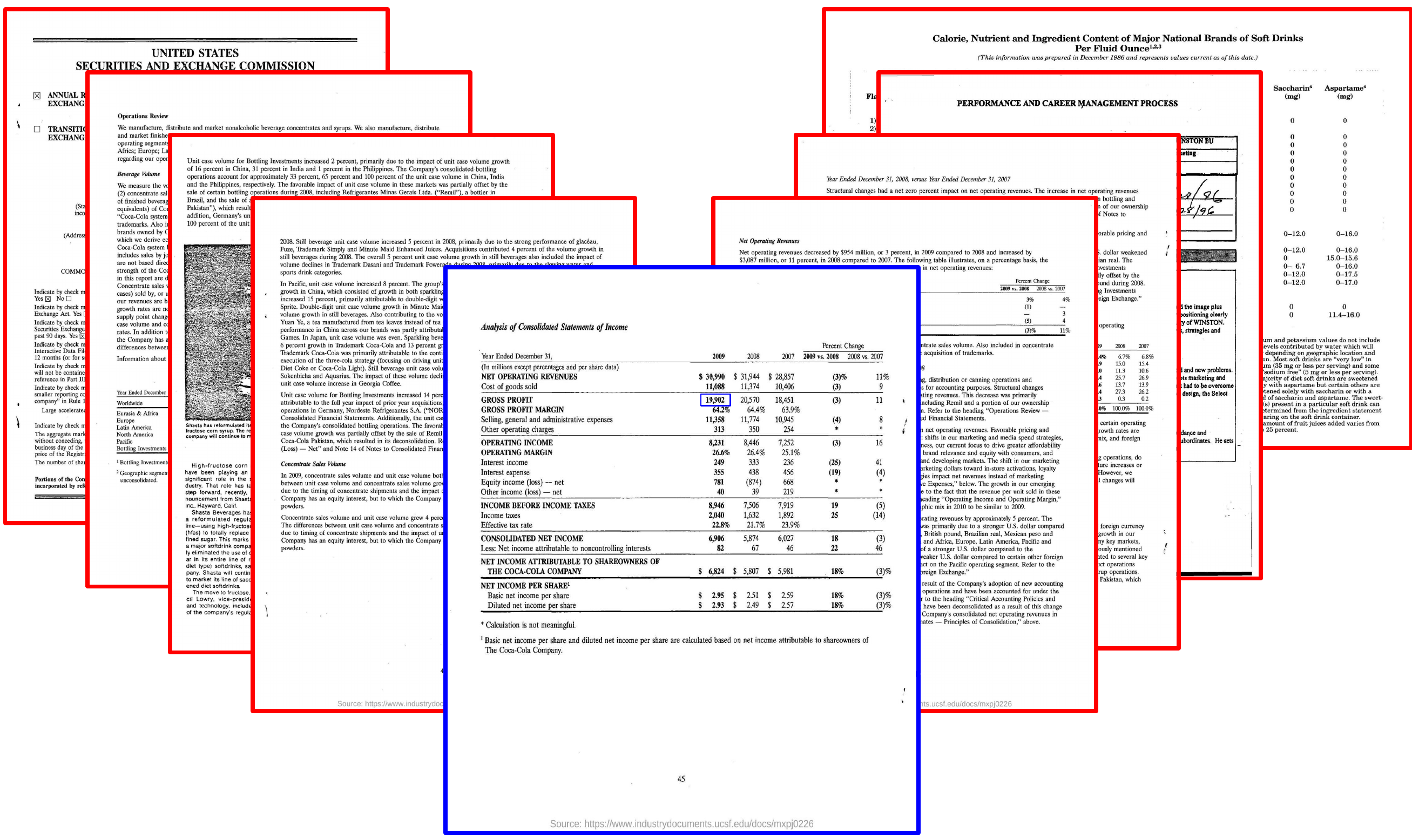}
        \textbf{Q:} What was the gross profit in the year 2009? \newline
        \textbf{A:} \$19,902
    \end{subfigure}
\caption{In the \textbf{MP-DocVQA task}, questions are posed over multi-page documents where methods are required to understand the text, layout and visual elements of each page in the document to identify the correct page (blue in the figure) and answer the question.}
\label{fig:task}
\vspace{-10pt}
\end{figure}

All existing datasets and methods for DocVQA focus on single page documents, which is far from real life scenarios. Documents are typically composed of multiple pages and therefore, in a real document management workflow all pages of a document need to be processed as a single set.

In this work we aim at extending single-page DocVQA to the more realistic \multipage setup. Consequently, 
we define a new task and propose a novel dataset, \datasetName, designed for Multi-Page Document Visual Question Answering. \datasetName is an extension of the SingleDocVQA~\cite{mathew2021docvqa} dataset where the questions are posed on documents with between 1 and 20 pages.

\begin{table*}[ht]
\small \centering
\begin{tabular}{lccccSSS}
\toprule
\multirow{2}{*}{\textbf{Dataset}} & \multirow{2}{*}{\textbf{Questions}} & \multirow{2}{*}{\textbf{Documents}} & \multirow{2}{*}{\textbf{Pages (Images)}} & \textbf{Avg. pages} &\textbf{Question}    & \textbf{Answer}      & \textbf{Document Avg.} \\
                                  &                                     &                                     &                                          & \textbf{per question} & \textbf{Avg. length} & \textbf{Avg. length} & \textbf{OCR Tokens}    \\
\midrule
SingleDocVQA~\cite{mathew2021docvqa}            & 50K  &   \hspace{9pt}6K   &   \hspace{5pt}12K     & 1.00      &  9.49    &   2.43      & 151.46    \\
VisualMRC~\cite{tanaka2021visualmrc}            & 30K  &   \hspace{5pt}10K  &   \hspace{5pt}10K     & 1.00      & 10.55   &   9.55      & 182.75    \\
InfographicsVQA~\cite{mathew2022infographicvqa} & 30K  &   \hspace{3pt}5.4K &   \hspace{3pt}5.4K    & 1.00      & 11.54   &   1.60      & 217.89    \\
DuReaderVis~\cite{qi2022dureadervis}            & 15K  &   158K             &   158K                & 1.3K   & 9.87       &   180.54    & 1968.21   \\
DocCVQA~\cite{tito2021document}                 & 20   &   \hspace{5pt}14K  &   \hspace{5pt}14K     & 14K      & 14.00   &   12.75     & 509.06    \\
TAT-DQA~\cite{zhu2022towards}                   & 16K  &   \hspace{3pt}2.7K &   \hspace{10pt}3K     & 1.07      & 12.54   &   3.44      & 550.27    \\
MP-DocVQA (ours)                                & 46K  &   \hspace{10pt}6K  &   \hspace{5pt}48K     & 8.27      & 9.90    &   2.20      & 2026.59   \\
\bottomrule
\end{tabular}
\vspace{-2pt}
\caption{Comparison between \datasetName and main DocVQA datasets.}
\vspace{-10pt}
\label{tab:datasets_stats}
\end{table*}

Dealing with multiple pages largely increases the amount of input data to be processed. This is particularly challenging for current state-of-the-art DocVQA methods \cite{xu2020layoutlm, xu2021layoutlmv2, huang2022layoutlmv3, powalski2021going} based on the Transformer architecture \cite{vaswani2017attention} that take as input textual, layout and visual features obtained from the words recognized by an OCR. As the complexity of the transformer scales up quadratically with the length of the input sequence, all these methods fix some limit on the number of input tokens which, for long \multipage documents, can lead to truncating a significant part of the input data. We will empirically show the limitations of current methods in this context. 

As an alternative, we propose the \methodFullName (\methodName), a \multimodal hierarchical encoder-decoder transformer build on top of T5~\cite{raffel2020exploring} which is capable to naturally process multiple pages by extending the input sequence length up to 20480 tokens without increasing the model complexity. In our architecture, the encoder processes separately each page of the document, providing a summary of the most relevant information conveyed by the page conditioned on the question. This information is encoded in a number of special \pageToken tokens, inspired in the \clsToken token of the BERT model \cite{devlin2018bert}. Subsequently, the decoder generates the final answer by taking as input the concatenation of all these summary \pageToken tokens for all pages. Furthermore, the model includes an additional head to predict the index of the page where the answer has been found. This can be used to locate the context of the answer within long documents, but also as a measure of explainability, following recent works in the literature ~\cite{wang2020general, tito2021document}. Correct page identification can be used as a way to distinguish which answers are the result of reasoning over the input data, and not dictated from model biases.

To summarize, the key contributions of our work are:
\vspace{-5pt}
\begin{enumerate}
    \item We introduce the novel dataset \datasetName containing questions over \multipage documents.
    \vspace{-8pt}
    \item We evaluate state-of-the-art methods on this new dataset and show their limitations when facing \multipage documents.
    \vspace{-8pt}
    \item We propose \methodName, a \multimodal hierarchical encoder-decoder method that can answer questions on \multipage documents and predict the page where the answer is found.
    \vspace{-8pt}
    \item We provide extensive experimentation to show the effectiveness of each component of our framework and explore the relation between the accuracy of the answer and the page identification result.
    \vspace{-4pt}
\end{enumerate}
The dataset, baselines and \methodName model code and weights are publicly available through the DocVQA Web portal\footnote{\href{https://rrc.cvc.uab.es/?ch=17}{\url{rrc.cvc.uab.es/?ch=17}}} and GitHub project\footnote{\href{https://github.com/rubenpt91/MP-DocVQA-Framework}{\url{github.com/rubenpt91/MP-DocVQA-Framework}}}.

\section{Related Work}
\vspace{-6pt}
\textbf{Document VQA datasets}:
DocVQA~\cite{mathew2020document, tito2021icdar} has seen numerous advances and new datasets have been released following the publication of the SingleDocVQA~\cite{mathew2021docvqa} dataset. This dataset consists of $50,000$ questions posed over industry document images, where the answer is always explicitly found in the text. The questions ask for information in tables, forms and paragraphs among others, becoming a high-level task that brought to classic DIAR algorithms an end purpose by conditionally interpreting the document images. 
Later on, InfographicsVQA~\cite{mathew2022infographicvqa} proposed questions on infographic images, with more visually rich elements and answers that can be either extractive from a set of multiple text spans in the image, a multiple choice given in the question, or the result of a discrete operation resulting in a numerical non-extractive answer. In parallel, VisualMRC~\cite{tanaka2021visualmrc} proposed open-domain questions on webpage screenshots with abstractive answers, which requires to generate longer answers not explicitly found in the text. 
DuReader\textsubscript{Vis}~\cite{qi2022dureadervis} is a Chinese dataset for open-domain document visual question answering, where the questions are queries from the Baidu search engine, and the images are screenshots of the webpages retrieved by the search engine results. Although the answers are extractive, $43\%$ of them are non-factual and much longer on average than the ones in previous DocVQA datasets. In addition, each image contains on average a bigger number of text instances.
However, due to the big size of the image collection, the task is posed as a 2-stage retrieval and answering tasks, where the methods must retrieve the correct page first, and answer the question in a second step. Similarly, the Document Collection Visual Question Answering (DocCVQA)~\cite{tito2021icdar} released a set of $20$ questions posed over a whole collection of $14,362$ single page document images. However, due to the limited number of questions and the low document variability, it is not possible to do training on this dataset and current approaches need to rely on training on SingleDocVQA. Finally, TAT-DQA~\cite{zhu2022towards} contains extractive and abstractive questions on modern financial reports. Despite that the documents might be \multipage, only 306 documents have actually more than one page, with a maximum of 3 pages.
Instead, our proposed \datasetName dataset is much bigger and diverse with $46,176$ questions posed over $5,928$ \multipage documents with its corresponding $47,952$ page images, which provides enough data for training and evaluating new methods on the new \multipage setting.

\begin{figure*}[ht]
    \centering
    \begin{subfigure}[b]{0.32\textwidth}
        \includegraphics[width=\textwidth]{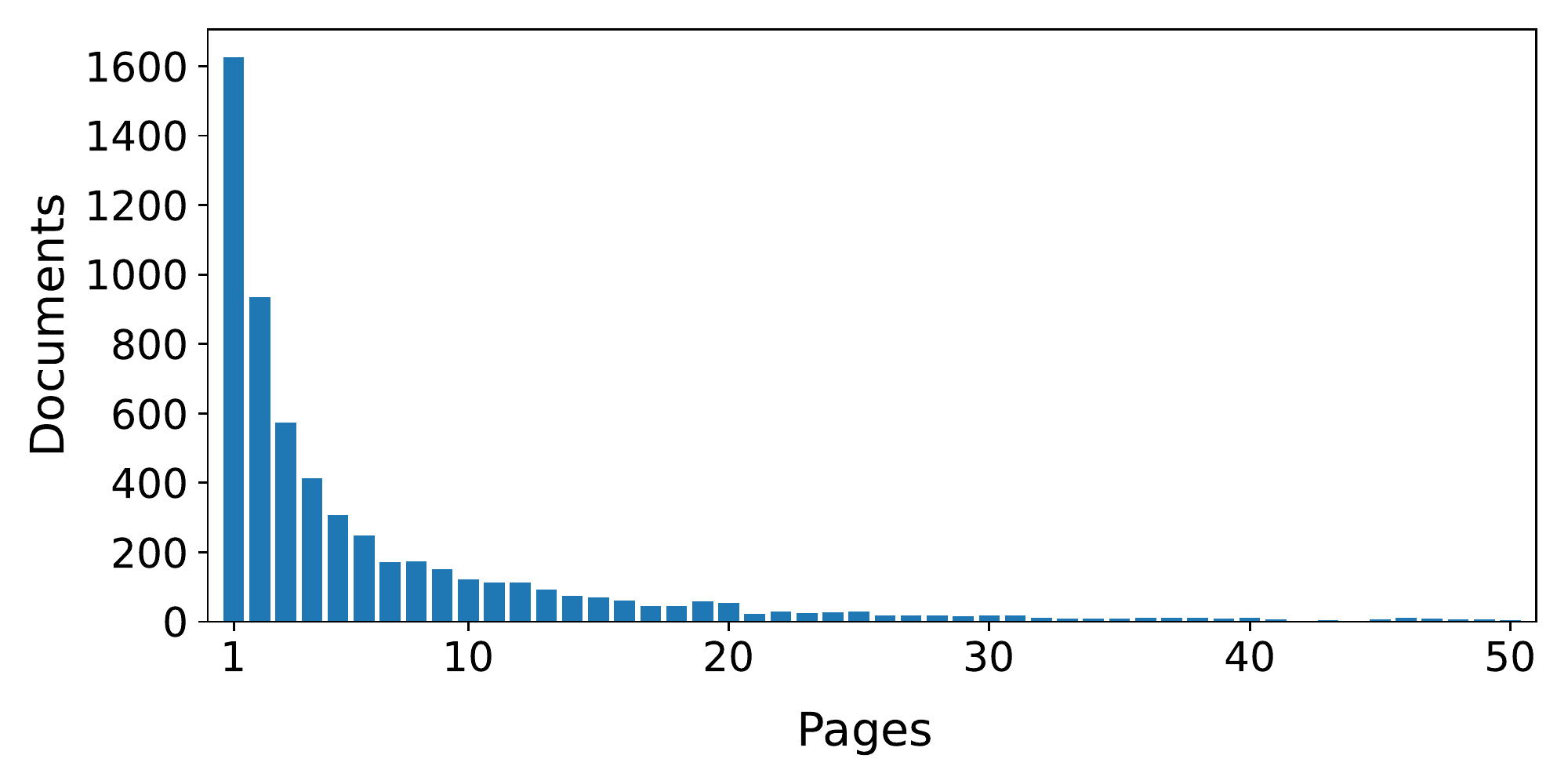}
        \caption{}
        \label{fig:doc_pages}
    \end{subfigure}
    \begin{subfigure}[b]{0.32\textwidth}
        \includegraphics[width=\textwidth]{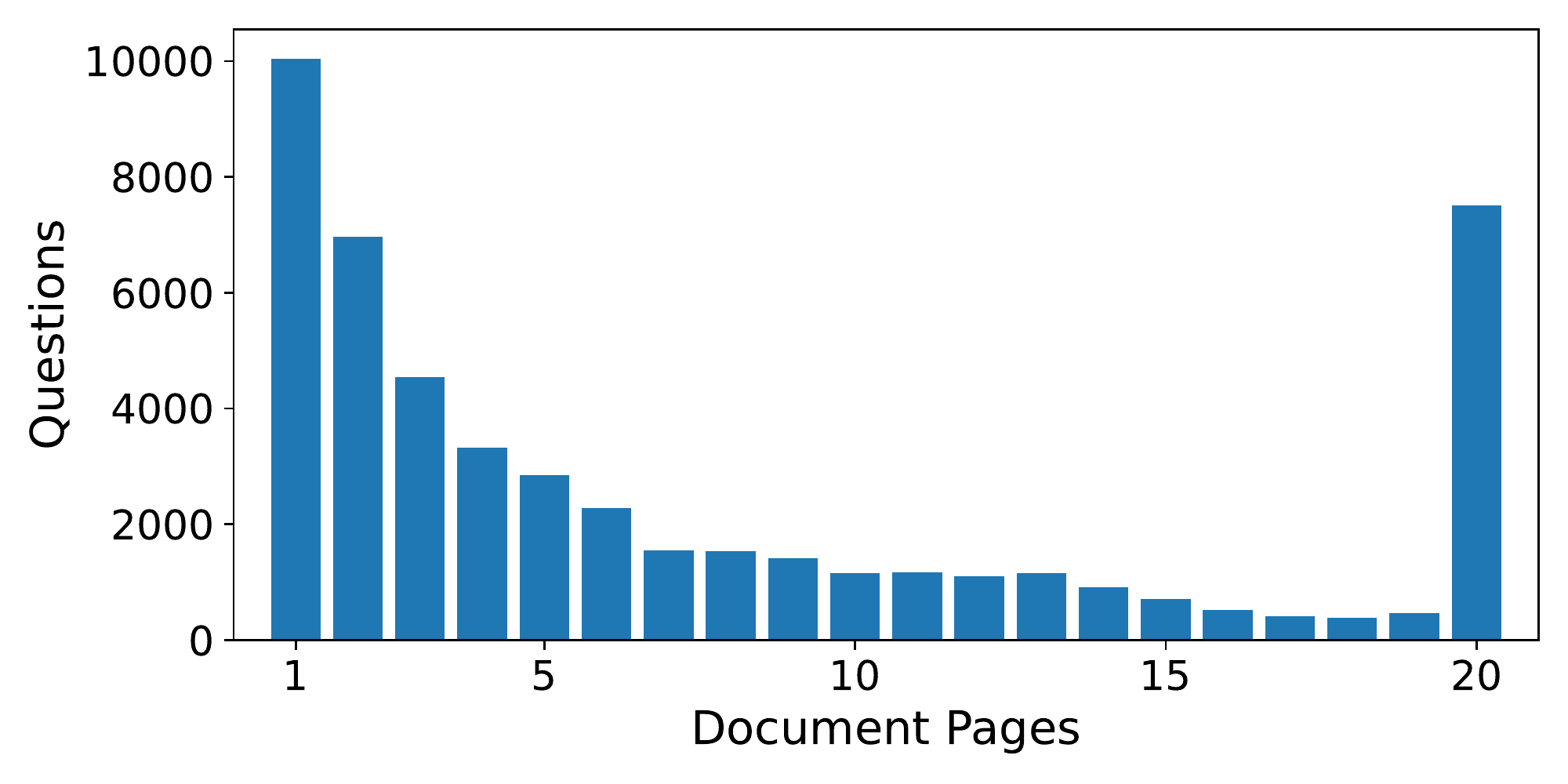}
        \caption{}
        \label{fig:questions_page_ranges}
    \end{subfigure}
    \begin{subfigure}[b]{0.32\textwidth}
        \includegraphics[width=\textwidth]{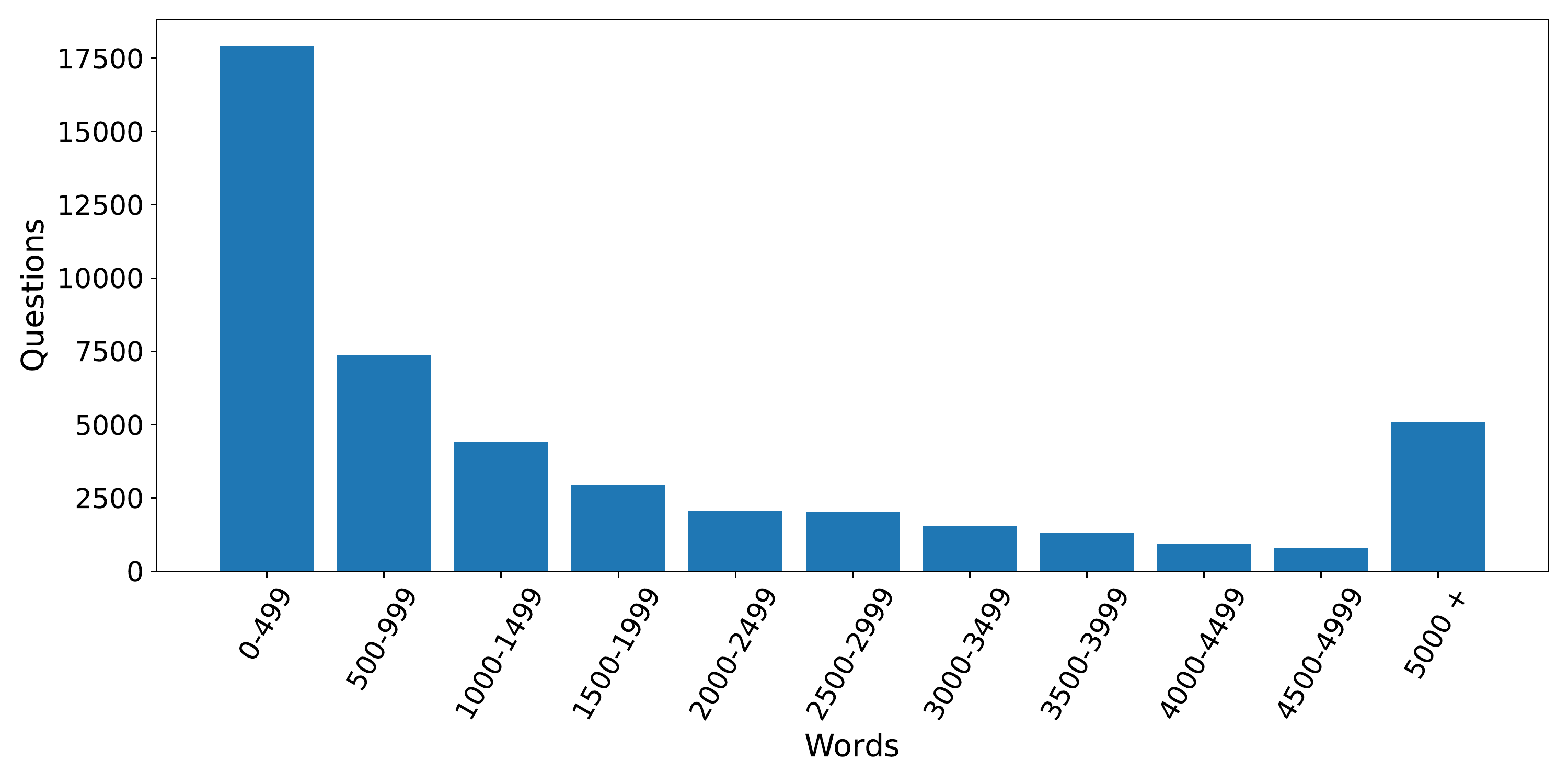}
        \caption{}
        \label{fig:words_per_question}
    \end{subfigure}
\caption{\textbf{MP-DocVQA statistics}. \textbf{(a)}: Distribution of the document length in term of pages of the documents included in \datasetName before applying the limit of 20 pages. \textbf{(b)}: Distribution of the document length in term of pages along the posed questions in the dataset. \textbf{(c)}: Number of recognized OCR words per question.}
\label{fig:dataset_statistics}
\end{figure*}


\textbf{Methods}: Since the release of the SingleDocVQA dataset, several methods have tackled this task from different perspectives. From NLP, Devlin \etal proposed BertQA~\cite{mathew2021docvqa} which consists of a BERT~\cite{devlin2018bert} architecture followed by a classification head that predicts the start and end indices of the answer span from the given context. While many models have extended BERT obtaining better results \cite{liu2019roberta, lan2019albert, garncarek2021lambert, sanh2019distilbert} by changing key hyperparameters during training or proposing new pre-training tasks, T5~\cite{raffel2020exploring} has become the backbone of many state-of-the-art methods~\cite{powalski2021going, biten2022latr, lu2022unified} on different NLP and \multimodal tasks. T5 relies on the original Transformer~\cite{vaswani2017attention} by performing minimal modifications on the architecture, but pre-training on the novel de-noising task on a vast amount of data. 

On the other hand, and specifically designed for document tasks, LayoutLM~\cite{xu2020layoutlm} extended BERT by decoupling the position embedding into 2 dimensions using the token bounding box from the OCR and fusing visual and textual features during the downstream task. Alternatively, LayoutLMv2~\cite{xu2021layoutlmv2} and TILT~\cite{powalski2021going}, included visual information into a \multimodal transformer and introduced a learnable bias into the self-attention scores to explicitly model relative position. In addition, TILT used a decoder to dynamically generate the answer instead of extracting it from the context. LayoutLMv3~\cite{huang2022layoutlmv3} extended its previous version by using visual patch embeddings instead of leveraging a CNN backbone and pre-training with 3 different objectives to align text, layout position and image context. In contrast, while all the previous methods utilize the text recognized with an off-the-shelf OCR, Donut~\cite{kim2022ocr} and Dessurt~\cite{davis2022end} are end-to-end encoder-decoder methods where the input is the document image along with  the question, and they implicitly learn to read as well as understand the semantics and layout of the images. 

However, the limited input sequence length of these methods make them unfeasible for tasks involving long documents such as the ones in \datasetName. Different methods\cite{dai2019transformer, beltagy2020longformer, zaheer2020big} have been proposed in the NLP domain to improve the modeling of long sequences without increasing the model complexity. 
\longformer~\cite{beltagy2020longformer} replaces the common self-attention used in transformers where each input attends to every other input by a combination of global and local attention. The global attention is used on the question tokens, which attend and are attended by all the rest of the question and context tokens, while a sliding window guides the local attention over the context tokens to attend the other locally close context tokens. While the standard self-attention has a complexity of $O(n^2)$, the new combination of global and local attention turns the complexity of the model into $O(n)$. Following this approach, \bigbird~\cite{zaheer2020big} also includes attention on randomly selected tokens that will attend and be attended by all the rest of the tokens in the sequence, which provides a better global representation while adding a marginal increase of the complexity in the attention pattern.

\section{\datasetName Dataset}

The Multi-Page DocVQA (\datasetName) dataset comprises 46K questions posed over 48K images of scanned pages that belong to 6K industry documents. The page images contain a rich amount of different layouts including forms, tables, lists, diagrams and pictures among others as well as text in handwritten, typewritten and printed fonts.

\subsection{Dataset creation} \label{subsec:dataset_creation}

Documents naturally follow a hierarchical structure where content is structured into blocks (sections, paragraphs, diagrams, tables) that convey different pieces of information. The information necessary to respond to a question more often than not lies in one relevant block, and is not spread over the whole document. This intuition was confirmed during our annotation process in this \multipage setting. The information required to answer the questions defined by the annotators was located in a specific place in the document. On the contrary, when we forced the annotators to use different pages as a source to answer the question, those become very unnatural and did not capture the essence of questions that we can find in the real world.

Consequently, we decided to use the SingleDocVQA~\cite{mathew2021docvqa} dataset, which already has very realistic questions defined on single pages. To create the new \datasetName dataset,
we took every image-question pair from SingleDocVQA~\cite{mathew2021docvqa} and added to every image the previous and posterior pages of the document downloaded from the original source UCSF-IDL\footnote{\url{https://www.industrydocuments.ucsf.edu/}}. As we show in \cref{fig:doc_pages} most of documents in the dataset have between $1$ and $20$ pages, followed by a long tail of documents with up to $793$ pages. We focused on the most common scenario and limited the number of pages in the dataset to $20$. For longer documents, we randomly selected a set of $20$ pages that included the page where the answer is found

Next, we had to analyze and filter the questions since we observed that some of the questions in the SingleDocVQA dataset became ambiguous when posed in a \multipage \setup (e.g. asking for the page number of the document). Consequently, we performed an analysis detailed in \cref{appendix:construction_details} to identify a set of key-words, such as \textit{`document'}, that when included in the text of the question, can lead to ambiguous answers in a \multipage setting, as they originally referred to a specific page and not to the whole \multipage document.


After removing ambiguous questions, the final dataset comprises $46,176$ questions posed over $47,952$ page images from $5,928$ documents. Notice that the dataset also includes documents with a single page when this is the case. 
Nevertheless, as we show in \cref{fig:questions_page_ranges}, the questions posed over \multipage documents represent the $85.95\%$ of the questions in the dataset. 

Finally, we split the dataset into train, validation and test sets keeping the same distribution as in SingleDocVQA. However, following this distribution some pages would appear in more than one split as they originate from the same document. To prevent this, we trim the number of pages used as context for such specific cases to ensure that no documents are repeated between training and validation/test splits.
In \cref{fig:questions_page_ranges} we show the number of questions according to the final document length.

To facilitate research and fair comparison between different methods on this dataset, along with the images and questions we also provide the OCR annotations extracted with Amazon Textract\footnote{\url{https://aws.amazon.com/textract/}} for all the \datasetPages document images (including page images beyond the $20$ page limit to not limit future research on longer documents).

\subsection{Dataset statistics}

As we show in \cref{tab:datasets_stats}, given that \datasetName is an extension of SingleDocVQA, the average question and answer lengths are very similar to this dataset in contrast to the long answers that can be found in the open-domain datasets VisualMRC and DuReader\textsubscript{Vis}. On the contrary, the main difference lies in the number of OCR tokens per document, which is even superior to the Chinese DuReader\textsubscript{Vis}. In addition, \datasetName adopts the \multipage concept, which means that not all documents have the same number of pages (\cref{fig:questions_page_ranges}), but also that each page of the document may contain a different content distribution, with varied text density, different layout and visual elements that raise unique challenges. Moreover, as we show in Figs. \ref{fig:questions_page_ranges} and \ref{fig:words_per_question} the variability between documents is high, with documents comprising between $1$ and $20$ pages, and between $1$ and $42,313$ recognized OCR words.

\begin{figure*}[ht]
  \vspace{-10pt}
  \includegraphics[width=0.90\textwidth]{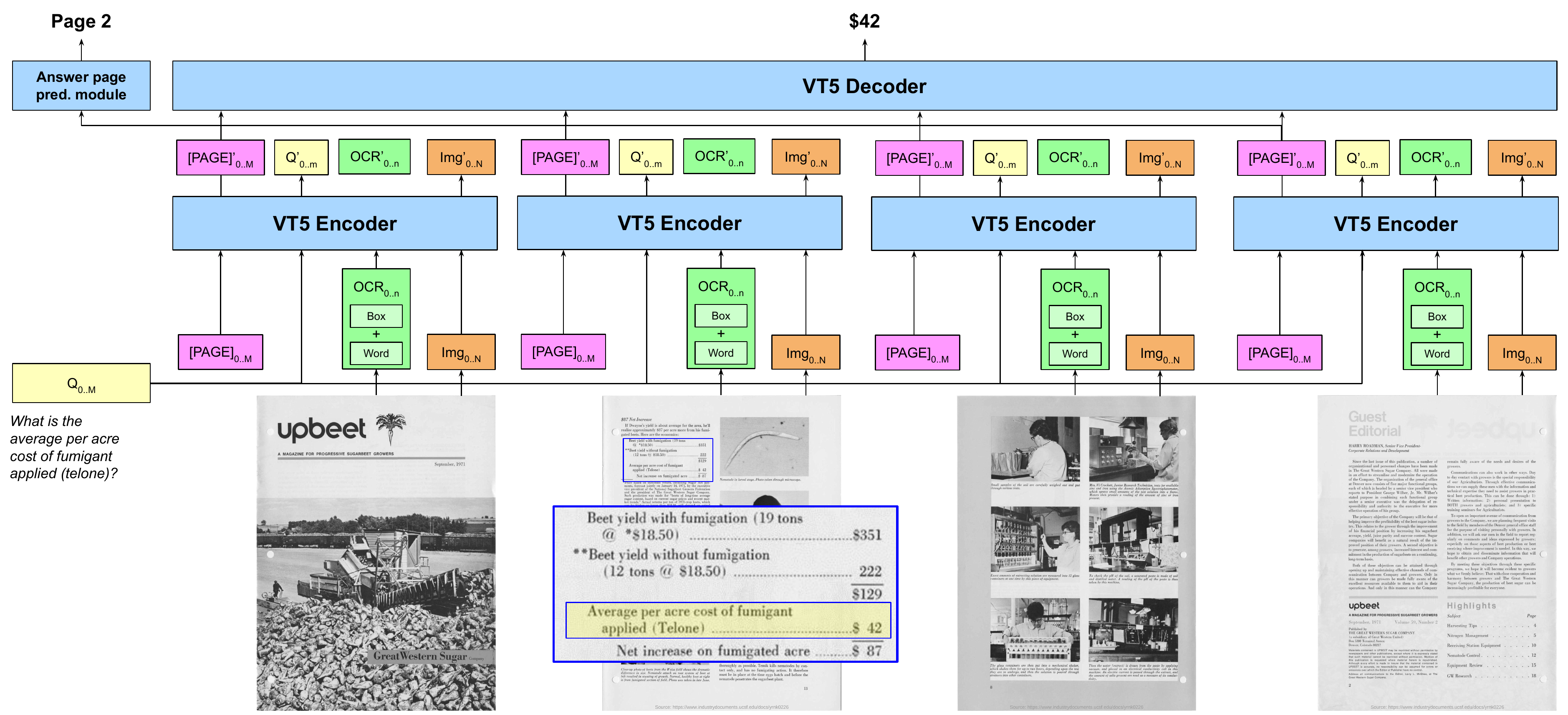}
  \vspace{-6pt}

  \caption{\textbf{Architecture of \methodName} model. The architecture is based on T5 with 2D layout features. Each page passes through the encoder to represent in the contextualized \pageTokenC tokens the most relevant information necessary to answer the posed question. Then, the \pageTokenC tokens of all pages are concatenated to provide the decoder with a holistic representation of the document at the time of generating the answer. In addition, a classification layer in the page answer page identification module outputs the page where the answer to the question is found, providing the model with an explainability measure of the answers which allows, among others, to understand if the answer has been inferred from the actual input data, or from a prior learned bias.}
  \label{fig:Hi-LT5}
  \vspace{-14pt}
\end{figure*}

\section{\methodName} \label{sec:method}
Although documents contain dense information, not all of them is necessary
to answer a given question. Following this idea, we propose the \methodFullName (\methodName), a hierarchical encoder-decoder \multimodal transformer where given a question, the encoder extracts the most relevant information from each page conditioned to the question and then, the decoder generates the answer from the summarized relevant information extracted from the encoder. Figure \ref{fig:Hi-LT5} shows an overview of the model. We can see that each page is independently processed by the encoder taking as input the sequence of OCR tokens (encoding both text semantics and layout features), a set of patch-based visual features and the encoded question tokens. In addition, a number of learnable \pageToken tokens are introduced to embed at the output of the encoder the summary of every page. These \pageToken tokens are concatenated and passed through the decoder to get the final answer. 
Moreover, in parallel to the answer generation, the answer page identification module predicts the page index where the information to answer the question is found, which can be used as a kind of explainability measure. We utilize the T5 architecture as the backbone for our method since the enormous amount of data and their novel de-noising task utilized during pretraining makes it an excellent candidate for the model initialization. In this section, we first describe each module, then how they are integrated and finally, the training process followed.



\textbf{Textual representation:} Following recent literature on document understanding~\cite{huang2022layoutlmv3, powalski2021going} which demonstrates the importance of layout information when working with Transformers, we utilize a spatial embedding to better align the layout information with the semantic representation. Formally, given an OCR token $O_{i}$, we define the associated word bounding box as $(x^{i}_{0}, y^{i}_{0}, x^{i}_{1}, y^{i}_{1})$. Following~\cite{biten2022latr}, to embed bounding box information, we use a lookup table for continuous encoding of one-hot vectors, and sum up all the spatial and semantic representations together:
\begin{equation}
    \small
    \mathcal{E}_{i} = E_{O} (O_{i}) + E_{x}(x^{i}_{0}) + E_{y}(y^{i}_{0})+E_{x}(x^{i}_{1}) + E_{y}(y^{i}_{1})
\label{eq:ocr_emb}
\end{equation}

\noindent{where $\mathcal{E}_{i}$ is the encoded representation for the OCR token $O_{i}$, and $E_{O}$, $E_{x}$ and $E_{y}$ are the learnable look-up tables. }

\textbf{Visual representation:} We leverage the Document Image Transformer (DIT)~\cite{li2022dit} pretrained on Document Intelligence tasks to represent the page image as a set of patch embeddings. Formally, given an image I with dimension $H \times W \times C$, is reshaped into $N$ 2D patches of size $P^{2} \times C$, where $(H, W)$ is the height and width, $C$ is the number of channels, $(P, P)$ is the resolution of each image patch, and $N = HW/P^{2}$ is the final number of patches. We map the flattened patches to $D$ dimensional space, feed them to DiT, pass the output sequence to a trainable linear projection layer and then feed it to the transformer encoder. We denote the final visual output as $V=\{v_{0}, \ldots, v_{N}\}$.

\textbf{\methodName hierarchical paradigm:} Inspired by the BERT~{\cite{devlin2018bert}} \clsToken token, which is used to represent the encoded sentence, we use a set of $M$ learnable \pageToken tokens to represent the page information required to answer the given question. Hence, we input the information from the different modalities along with the question and the learnable tokens to the encoder to represent in the \pageToken tokens the most relevant information of the page conditioned by the question. More formally, for each page $p_{j} \in P=\{p_{0}, \ldots, p_{K}\}$, let $V_{j}=\{v_{0}, \ldots, v_{N}\}$ be the patch visual features, $Q=\{q_{0}, \ldots, q_{m}\}$ the tokenized question, $O_{j}=\{o_{1}, \ldots, o_{n}\}$ the page OCR tokens and $K_{j}=\{k_{0}, \ldots, k_{M}\}$ the learnable \pageToken tokens. Then, we embed the OCR tokens and question using \cref{eq:ocr_emb} to obtain the OCR $\mathcal{E}_{j}^{o}$ and question $\mathcal{E}^{q}$ encoded features. And concatenate all the inputs $[K_{j};V_{j};\mathcal{E}^{q};\mathcal{E}_{j}^{o}]$ to feed to the transformer encoder. Finally, all the contextualized $K^{'}$ output tokens of all pages are concatenated to create a holistic representation of the document $D=[K_{0}^{'}; \ldots; K_{K}{'}]$,
which is sent to the decoder that will generate the answer, and to the answer page prediction module.

\begin{table*}[ht]
\centering
\small
\vspace{-10pt}

\begin{tabular}{lccccccc}
\toprule
    & \multicolumn{1}{l}{}      & \multicolumn{1}{l}{}  & \textbf{Max Seq.} & &  &   & \textbf{Ans. Page}    \\
    
\multirow{-2}{*}{\textbf{Model}} & \multicolumn{1}{l}{\multirow{-2}{*}{\textbf{Size}}} & \multicolumn{1}{l}{\multirow{-2}{*}{\textbf{Parameters}}} & \textbf{Length}        & \multirow{-2}{*}{\textbf{Setup}}      & \multirow{-2}{*}{\textbf{Accuracy}} & \multirow{-2}{*}{\textbf{ANLS}} & \textbf{Accuracy}                   \\

\midrule
                            &       &       &       & Oracle    & 39.77     & 0.5904    & 100.00    \\
BERT~\cite{devlin2018bert}  & Large & 334M  & 512   & \MaxConf    & 34.78     & 0.5347    &  \hspace{5pt}71.24    \\
                            &       &       &       & Concat    & 27.41     & 0.4183    &  \hspace{5pt}51.61    \\
\hdashline
                                         &      &       &       & Oracle    & 52.48     & 0.6177    & 100.00    \\
\longformer~\cite{beltagy2020longformer} & Base & 148M  & 4096  & \MaxConf    & 45.87     & 0.5506    &  \hspace{5pt}70.37    \\
                                         &      &       &       & Concat    & 43.91     & 0.5287    &  \hspace{5pt}71.17    \\
\hdashline
                                &       &       &       & Oracle    & 55.31     & 0.6450    & 100.00    \\
\bigbird~\cite{zaheer2020big}   & Base  & 131M  &  4096 & \MaxConf    & \textbf{49.57}     & 0.5854    & \hspace{5pt}72.27     \\
                                &       &       &       & Concat    & 41.06     & 0.4929    & \hspace{5pt}67.54     \\
\hdashline
                                        &       &       &       & Oracle    & 58.81     & 0.6729    & 100.00 \\
LayoutLMv3~\cite{huang2022layoutlmv3}   & Base  & 125M  & 512   & \MaxConf    & 42.70     & 0.5513    & 74.02  \\  
                                        &       &       &       & Concat    & 38.47     & 0.4538    & 51.94  \\
\hdashline
                                &       &       &       & Oracle    & \textbf{59.00}     & \textbf{0.6814}    & 100.00    \\
T5~\cite{raffel2020exploring}   & Base  & 223M  & 512   & \MaxConf    & 32.68     & 0.4028    & \hspace{5pt}46.05    \\
                                &       &       &       & Concat    & 41.80     & 0.5050    & \hspace{20pt}-- \\
\hdashline
\multirow{-0.5}{*}{\methodName (Ours)} & \multirow{-0.5}{*}{Base} & \multirow{-0.5}{*}{316M} & \multirow{-0.5}{*}{20480} & Oracle & 50.01 & 0.6572 & 100.00 \\
 &  &  &  & Multipage & 48.28 & \textbf{0.6201} & \hspace{5pt}\textbf{79.23} \\

\bottomrule
\end{tabular}
\vspace{-1.5mm}
\captionsetup{width=.87\textwidth}
\caption{\textbf{Baselines and proposed method \methodName results on \datasetName dataset}. Baselines are evaluated on three different setups: oracle, concat and \logits. The proposed method is evaluated only on the oracle setup and the realistic \multipage setting. We highlight in bold the best results for the oracle and any multi-page (oracle and \logits) setup.}
\vspace{-3mm}

\label{tab:methods_results}
\end{table*}

\textbf{Answer page identification module}: Following the trend to look for interpretability of the answers in VQA~\cite{wang2020general}, in parallel to the the answer generation in the decoder, the contextualized \pageToken tokens $D$ are fed to a classification layer that outputs the index of the page where the answer is found.

\textbf{Pre-training strategy:}
Since T5 was trained without layout information, inspired by \cite{biten2022latr} we propose a hierarchical layout-aware pretraining task to align the layout and semantic textual representations, while providing the \pageToken tokens with the ability to attend to the other tokens. Similar to the standard de-noising task, the layout-aware de-noising task masks a span of tokens and forces the model to predict the masked tokens. Unlike the normal de-noising task, the encoder has access to the rough location of the masked tokens, which encourages the model to fully utilize the layout information when performing this task. In addition, the masked tokens must be generated from the contextualized $K^{'}$ \pageToken tokens created by the encoder, which forces the model to embed the tokens with relevant information regarding the proposed task.

\textbf{Training strategy:} Even though \methodName keeps the same model complexity as the sum of their independent components (T5\textsubscript{BASE} (223M) + DiT\textsubscript{BASE} (85M)) and despite being capable to accept input sequences of up to 20480 tokens, the amount of gradients computed at training time scales linearly with the number of pages since each page is passed separately through the encoder and the gradients are stored in memory. Consequently, it is similar to have a batch size $P$ times bigger in the encoder compared to a single page setting. While this could be tackled by parallelizing the gradients corresponding to a set of pages into different GPUs, we offer an alternative strategy using limited resources. We train the model on shortened versions of the documents with only two pages: the page where the answer is found and the previous or posterior page. Even though this drops the overall performance of the model, as we show in \cref{appendix:train_doc_pages}, training with only 2 pages is enough to learn the hierarchical representation of the model achieving results close to the ones using the whole document, and offers a good trade-off in terms of memory requirements. However, after the training phase the decoder and the answer page identification module can't deal with the full version of the documents of up to 20 pages. For this reason, we perform a final fine-tuning phase using the full-length documents and freezing the encoder weights.

\section{Experiments} \label{sec:experiments}

To evaluate the performance of the methods, we use the standard evaluation metrics in DocVQA, accuracy and Average Normalized Levenshtein Similarity (ANLS)~\cite{biten2019scene}. To assess the page identification we use accuracy.

\begin{figure*}[ht]
  \centering
  \vspace{-10pt}
  \includegraphics[width=\textwidth]{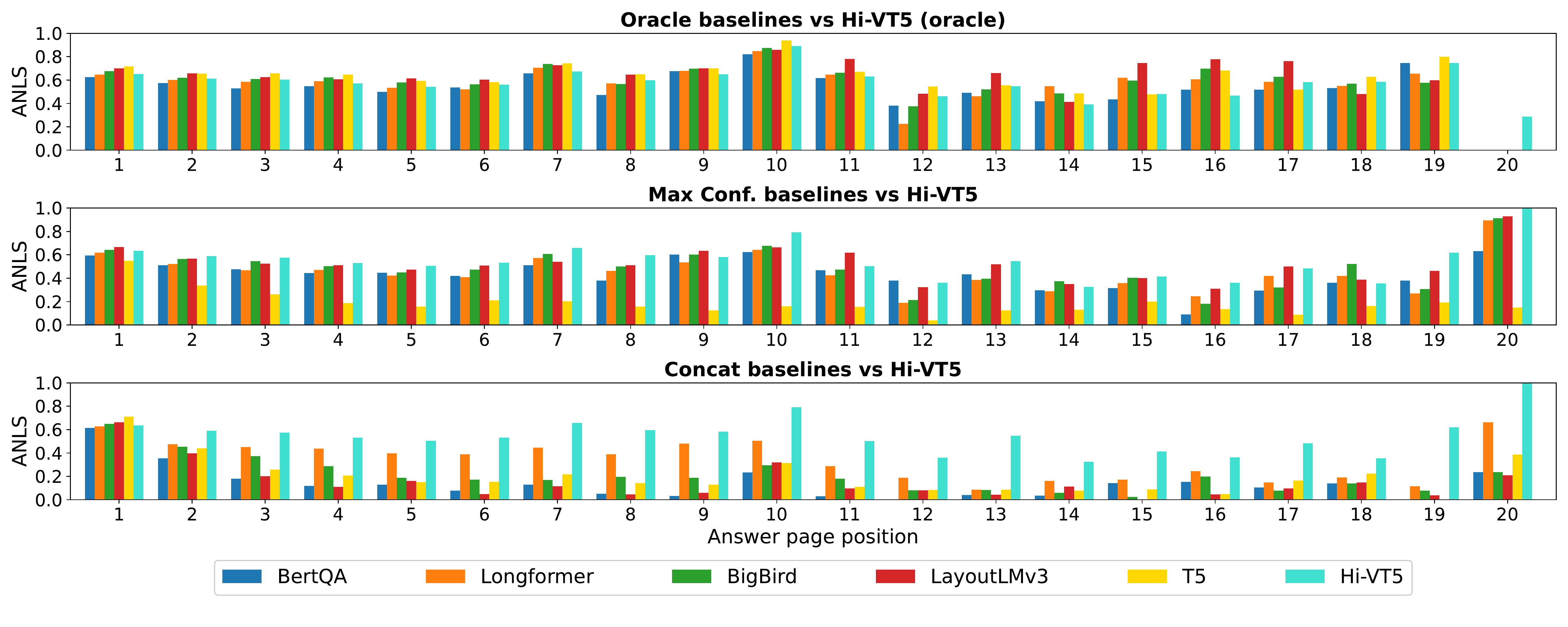}
  \vspace{-20pt}
  \caption{\textbf{Methods ANLS by answer page position}. The figure shows the answering performance of the different baselines and \methodName in the oracle \setup (top), and the baselines in the \logits (middle) and concat (bottom) \setup against \methodName using its answer page identification module. Notice that the breakdown of the scores is NOT performed on the number of the document pages, but in which page the answer is found.}
  \label{fig:methods_anls_by_answer_page}
  \vspace{-10pt}
\end{figure*}

\subsection{Baselines} 

As Multi-Page DocVQA is a new task, we adapt several state-of-the-art methods as baselines to analyze their limitations in the \multipage setup and compare their performance against our proposed method. We choose BERT~\cite{devlin2018bert} because it was the first question-answering method based on transformers, and it shows the performance of such a simple baseline. \longformer~\cite{beltagy2020longformer} and \bigbird~\cite{zaheer2020big} because they are specially designed to deal with long sequences, which might be beneficial for the \multipage setting. In the case of \bigbird it can work following two different strategies. The former, Internal Transformer Construction (ITC) only sets the global attention over one single token, while the Extended Transformer Construction (ETC) sets the global attention over a set of tokens. Although the latter strategy is the desired \setup for question-answering tasks by setting all the question tokens with global attention, the current released code only supports the ITC strategy and hence, we limit our experiments to this attention strategy. We also use LayoutLMv3~\cite{huang2022layoutlmv3} because it is the current public state-of-the-art method on the SingleDocVQA task and uses explicit visual features by representing the document in image patches. Finally, T5~\cite{raffel2020exploring} because it is the only generative baseline and the backbone of our proposed method.

However, all these methods are not directly applicable to a \multipage scenario. Consequently, we define three different \setups to allow them to be evaluated on this task. In the \oracle \setup, only the page that contains the answer is given as input to the transformer model. Thus, this setup aims at mimicking the Single page DocVQA task. It shows the raw answering capabilities of each model regardless of the size of the input sequences they can accept. So, it should be seen as a theoretical maximum performance, assuming that the method has correctly identified the page where the information is found. In the \concat \setup, the context input to the transformer model is the concatenation of the contexts of all the pages of the document. This can be considered the most realistic scenario where the whole document is given as a single input. It is expected that the large amount of input data becomes challenging for the baselines. The page corresponding to the predicted start index is used as the predicted page, except for T5, since being a generative method it does not predict the start index. 
Finally, \logitsx is the third \setup, which is inspired in the strategy that the best performing methods in the DocCVQA challenge \cite{tito2021document} use to tackle the big collection of documents. 
In this case, each page is processed separately by the model, providing an answer for every page along with a confidence score in the form of logits. Then, the answer with the highest confidence is selected as the final answer with the corresponding page as the predicted answer page.

For BERT, \longformer, \bigbird and T5 baselines we create the context following the standard practice of concatenating the OCR words in the image following the reading (top-left to bottom-right) order. For all the methods, we use the Huggingface~\cite{wolf2020transformers} implementation and pre-trained weights from the most similar task available. We describe the specific initialization weights and training hyperparameters in \cref{appendix:hyperparameters}.

\subsection{Baseline results}


As we show in \cref{tab:methods_results}, the method with the best answering performance in the oracle setup (i.e. when the answer page is provided) is T5, followed by LayoutLMv3, \bigbird, \longformer and BERT. This result is expected since this \setup is equivalent to the single page document setting, where T5 has already demonstrated its superior results. In contrast, in the \logits \setup, when the logits of the model are used as a confidence score to rank the answers generated for each page, T5 performs the worst because the softmax layer used across the vocabulary turns the logits unusable as a confidence to rank the answers. 
Finally, in the concat \setup, when the context of all pages are concatenated \longformer outperforms the rest, showing its capability to deal with long sequences as seen in \cref{fig:methods_anls_by_answer_page}, which shows that the performance gap increases as long as the answer page is placed at the end of the document. The second best performing method in this setting is T5, which might seem surprising due to its reduced sequence length. However, looking at \cref{fig:methods_anls_by_answer_page} it is possible to see that is good on questions whose answers can fit into the input sequence, while it is not capable to answer the rest. In contrast, \bigbird is capable to answer questions that require long sequences since its maximum input length is 4096 as \longformer. Nevertheless, it performs worse due to the ITC strategy \bigbird is using, which do not set global attention to all question tokens and consequently, as long as the question and the answer tokens become more distant, it is more difficult to model the attention between the required information to answer the question.

\subsection{\methodName results}

In our experiments we fixed the number of \pageToken tokens to $M=10$, through experimental validation explained in detail in \cref{appendix:num_page_tokens}. We observed no significant improvements beyond this number. 
We pretrain \methodName on hierarchical aware de-noising task on a subset of 200,000 pages of OCR-IDL~\cite{biten2022ocr} for one epoch. Then, we Train on \datasetName for 10 epochs with the 2-page shortened version of the documents and finally, perform the fine-tuning of the decoder and answer page identification module with the full length version of the documents for 1 epoch. During training and fine-tuning all layers of the DiT visual encoder are frozen except a  last fully connected projection layer.

\methodName outperforms all the other methods both on answering and page identification in the concat and \logits setups, which are the most realistic scenarios. In addition, when looking closer at the ANLS per answer page position (see \cref{fig:methods_anls_by_answer_page}), the performance gap becomes more significant when the answers are located at the end of the document, even compared with \longformer, which is specifically designed for long input sequences. In contrast, \methodName 
shows a performance drop in the \oracle setup compared to the original T5. This is because it must infer the answer from a compact summarized representation of the page, while T5 has access to the whole page representation. This shows that the page representation obtained by the encoder has still margin for improvement. 

\begin{figure}[ht]
  \includegraphics[width=1.0\linewidth]{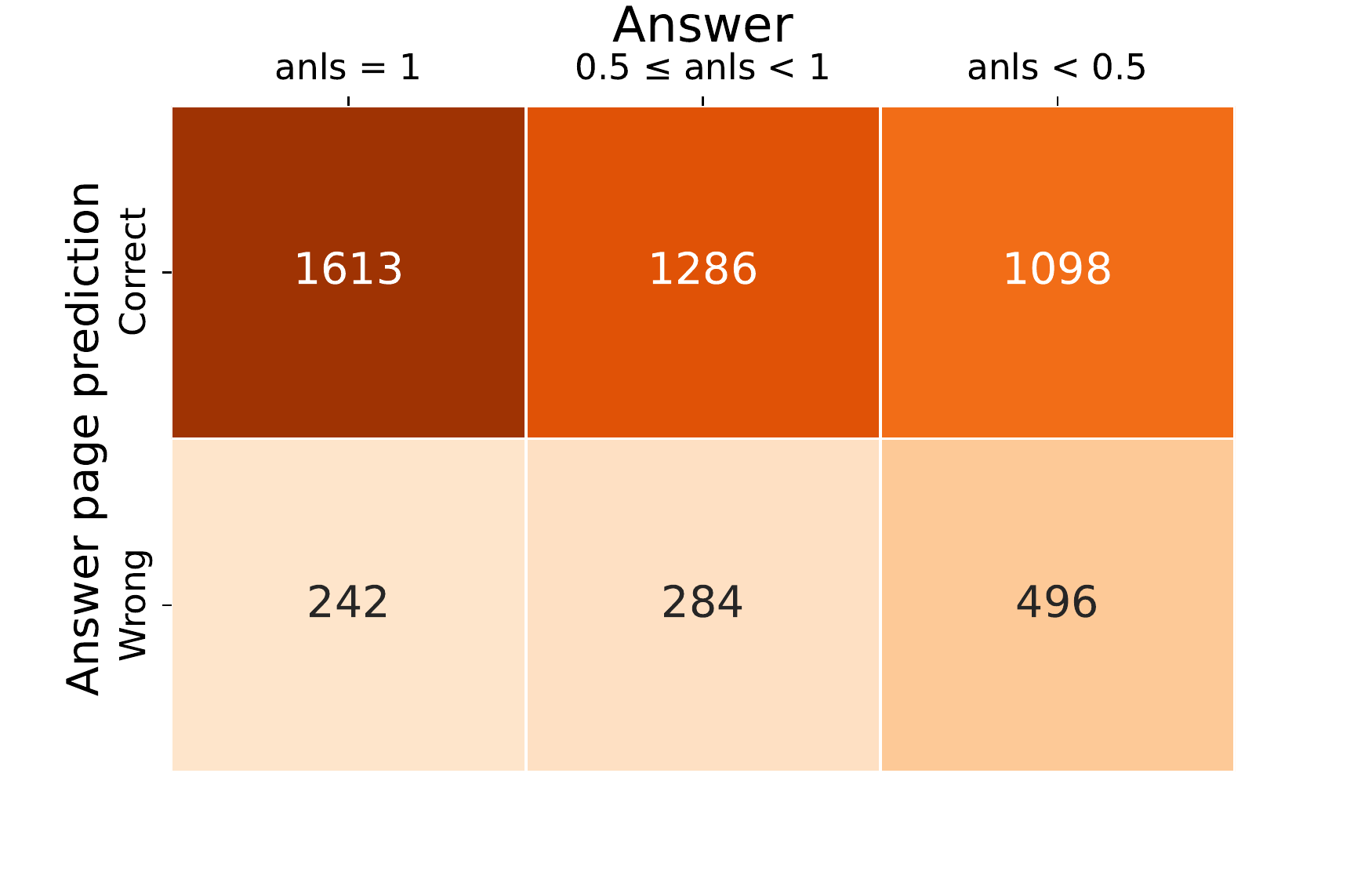}
  \vspace{-11mm}
  \caption{Matrix showing the \methodName correct and wrong answered questions depending on the answer page prediction module result.}
  \label{fig:ret_answ_matrix}
\end{figure}

Finally, identifying the page where the answer is found at the same time as answering the question allows to better interpret the method's results. In~\cref{tab:methods_results} we can see that \methodName obtains a better answer page identification performance than all the other baseline methods. In addition, in \cref{fig:ret_answ_matrix} we show that it is capable to predict the correct page even when it cannot provide the correct answer. Interestingly, it answers correctly some questions for which the predicted page is wrong, which means that the answer has been inferred from a prior learned bias instead of the actual input data. We provide more details by analyzing the attention of \methodName in \cref{appendix:attention_viz}.

\vspace{-6pt}

\section{Ablation studies} \label{sec:ablation}

To validate the effectiveness of each feature proposed in \methodName, we perform an ablation study and show results in \cref{tab:ablation_results}.
Without the answer page prediction module the model performs slightly worse on the answering task, showing that both tasks are complementary and the correct page prediction helps to answer the question. The most significant boost comes from the hierarchical de-noising pre-training task, since it allows the \pageToken tokens to learn better how to represent the content of the document. The last fine-tuning phase where the decoder and the answer page prediction module are adapted to the 20 pages maximum length of the \datasetName documents, is specially important for the answer page prediction module because the classification layer predicts only page indexes seen during training and hence, without finetuning it can only predict the first or the second page of the documents as the answer page. Finally, when removing the visual features the final scores are slightly worse, which has also been show in other works in the literature~\cite{huang2022layoutlmv3, biten2022latr, powalski2021going}, the most relevant information is conveyed within the text and its position, while explicit visual features are not specially useful for grayscale documents.

\begin{table}[ht]
\renewcommand*{\arraystretch}{1.3}
\vspace{-6pt}
\small
\centering
\begin{tabular}{lccc}
\toprule
\textbf{Method} & \textbf{Accuracy} & \textbf{ANLS} & \textbf{Ans. Page Acc.} \\
\midrule
Hi-VT5                & 48.28 & 0.6201 & 79.23  \\
\quad --2D-pos        & 46.12 & 0.5891 & 78.21  \\
\quad --Vis. Feat.    & 46.82 & 0.5999 & 78.22  \\
\quad --APPM          & 47.78 & 0.6130 & 00.00  \\
\quad --Pretrain      & 42.10 & 0.5864 & 81.47  \\
\quad --Fine-tune     & 42.86 & 0.6263 & 55.74  \\
\bottomrule
\end{tabular}
\vspace{-1.5mm}
\caption{\textbf{\methodName ablation studies}. We study the effect of removing different components independently from \methodName namely the 2D position embedding (2D-pos), visual features (Vis. Feat.), the answer page prediction module (APPM), the pretraining (Pretrain) and the last 
fine-tuning (Fine-tune) phase of the decoder and answer page prediction module.}
\label{tab:ablation_results}
\end{table}

\vspace{-6mm}
\section{Conclusions} \label{sec:conclusions}

In this work, we propose the task of Visual Question Answering on \multipage documents and make public the \datasetName dataset. To show the challenges the task poses to current DocVQA methods, we convey an analysis of state-of-the-art methods showing that even the ones designed to accept long sequences are not capable to answer questions posed on the final pages of a document. In order to address these limitations, we propose the new method \methodName that, without increasing the model complexity, can accept sequences up to 20,480 tokens and answer the questions regardless of the page in which the answer is placed. Finally, we show the effectiveness of each of the components in the method, and perform an analysis of the results showing how the answer page prediction module can help to identify answers that might be inferred from prior learned bias instead of the actual input data. 

\section*{Acknowledgements}

This work has been supported by the UAB PIF scholarship B18P0070, the Consolidated Research Group 2017-SGR-1783 from the Research and University Department of the Catalan Government, and the project PID2020-116298GB-I00, from the Spanish Ministry of Science and Innovation.

{\small
\bibliographystyle{ieee_fullname}
\bibliography{references}
}

\newpage ~ \newpage
\appendix

\section{\datasetName construction process}  \label{appendix:construction_details}

As described in \cref{subsec:dataset_creation}, the source data of the \datasetName dataset is the SingleDocVQA~\cite{mathew2021docvqa} dataset. The first row of \cref{tab:construction_process_stats} shows the number of documents, pages and questions in this dataset. The first step to create the \datasetName dataset was to download and append to the existing documents their previous and posterior pages, increasing the number of page images from 12,767 to 64,057, as shown in the second row of \cref{tab:construction_process_stats}.

\begin{table}[H]
\resizebox{\columnwidth}{!}{
\begin{tabular}{lcccc}
\toprule
                       & Documents & Pages  & Questions \\
\midrule
SingleDocVQA              & 6,071     & 12,767 & 50,000 \\
\datasetName (full)       & 6,071     & 64,057 & 50,000 \\
\datasetName (filtered)   & 5,928     & 60,884 & 46,176 \\ \datasetName (20 page limit) & 5,928  & 47,952 & 46,176 \\
\datasetName (multi-page) & 3,824     & 39,688 & 39,688 \\   
\bottomrule
\end{tabular}}
\caption{Statistics of the \datasetName during its construction process.}
\label{tab:construction_process_stats}
\end{table}

However, not all questions are suited to be asked on multi-page documents. Therefore, we performed an analysis based on manually selected key-words that appear in the questions, searching for those questions whose answer becomes ambiguous when they are posed over a \multipage document. Some of the selected key-words are shown in table \cref{tab:key-word_analysis}, along with some examples of potentially ambiguous questions containing those key-words. The most clear example is with the word 'document'. When looking at each document page separately, we can observe that many times they start with a big text on the top that can be considered as the title, which is actually the answer in the single page DocVQA scenario when the question asks about the title of the document. However, this pattern is repeated in every page of the document, making the question impossible to answer when multiple pages are taken into account. Moreover, even if there is only one page with a title, the answer can still be considered wrong, since the title of the document is always found in the first page like in the example in \cref{fig:task}. On the other hand, when we analyzed more closely  other potentially ambiguous selected key-words such as 'image', 'appears' or 'graphic' we found out that the answers were not always ambiguous and also the amount of questions with those words was negligible compared to the entire dataset. Thus, we decided to keep those questions in our dataset. Finally, we found that the key-word 'title' was mostly ambiguous only when it was written along with the word 'document'. Hence, we decided to remove only the questions with the word 'document' in it, while keeping all the rest. This filtered version, which is represented in the third row of \cref{tab:construction_process_stats} is the dataset version that was released and used in the experiments.

Nevertheless, it is important to notice that not all the questions in \datasetName are posed over multi-page documents. We keep the documents with a single page because they are also a possible case in a real life scenario. However, as showed in the fourth row of \cref{tab:construction_process_stats}, the questions posed over multiple pages represent the 85.95\% of all the questions in the dataset.

\section{Number of \pageToken tokens} \label{appendix:num_page_tokens}

\methodName embeds the most relevant information from each page conditioned by a question into $M$ \pageToken tokens. However, we hypothesize that contrary to BERT~\cite{devlin2018bert}, which represents a sentence with a single \clsToken token, \methodName will require more than one token to represent a whole page, since it conveys more information. Consequently, we perform an experimental study to find the optimum number of \pageToken tokens to use. We start by defining the maximum number of tokens $M$ that can be used, which is limited by the decoder input sequence length $S$, and the number of pages $P$ that must be processed. Formally, 
\vspace{-2mm}
\begin{equation} M=int\left(\frac{S}{P}\right) \label{eq:page_tokens_tradeoff}
\vspace{-2mm}
\end{equation}
We can set $M$ as an hyperparameter to select depending on the number of pages we need to process, 
where in the extreme cases we can represent a single page with 1024 \pageToken tokens, or a 1024 page document with a single token for each page.

Constraining to the 20 pages documents scenario of \datasetName, the maximum possible number of tokens $M$ would be 51. We performed a set of experiments with different \pageToken tokens to find the optimal value. As we show in \cref{tab:page_tokens_exp}, the model is able to answer correctly some questions even when using only one or two tokens. However, the performance increases significantly when more tokens are used. Nevertheless, the model does not benefit from using more than 10 tokens, since it performs similarly either with 10 or 25 tokens. Moreover, the performance decreases when using more. This can be explained because the information extracted from each page can be fully represented by 10 tokens, while using more, not only does not provide any benefit, but also makes the training process harder.

\begin{table}[ht]
\centering
\small
\begin{tabular}{cccc}
\toprule
\textbf{\pageToken} & \multirow{2}{*}{\textbf{Accuracy}} & \multirow{2}{*}{\textbf{ANLS}} & \textbf{Ans. Page} \\
\textbf{Tokens} &                           &                       & \textbf{Accuracy}  \\
\midrule
1                 & 36.41    & 0.4876 & 79.87  \\
2                 & 37.94    & 0.5282 & 79.88  \\
5                 & 39.31    & 0.5622 & 80.77  \\
10                & 42.10    & 0.5864 & 81.47  \\
25                & 42.16    & 0.5896 & 81.35  \\
50                & 30.63    & 0.5768 & 59.18  \\         
\bottomrule
\end{tabular}
\caption{Results of \methodName with different \pageToken tokens.}
\label{tab:page_tokens_exp}
\end{table}

\begin{table*}[ht]
\resizebox{\textwidth}{!}{
\begin{tabular}{p{0.25\linewidth}p{0.25\linewidth}p{0.25\linewidth}p{0.25\linewidth}} 
\toprule
{\textbf{Document} (3824)}     & \textbf{Image} (72)        & \textbf{Appears} (15)      & \textbf{Title} (1836)      \\
\midrule
What is the subject of the \textbf{document}/letter?          &
What is the number of calories written in the \textbf{image}? &
Whose name \textbf{appears} on top of the schedule?           &
What is the \textbf{title} of this document?                  \\
  &  &  &  \\
  
What is the title of the \textbf{document}?                             &
What does the \textbf{image} say?                                       &
What is the name of registered agent as it \textbf{appears} of record?  &
What is the \textbf{title} of the table?                                \\
  &  &  &  \\

What date is the meeting scheduled to develop the overall structure of the \textbf{document}?   &
In the \textbf{image} of the man with a trophy, what is the name of the awards given?           &
Who \textbf{appears} in the photograph at the top of the document standing alone with Nehru?    &
Which are prescribed earlier in the treatment of type 2 diabetes under the \textbf{title} of "critical success factors"?               \\
  &  &  &  \\

What is the subject of the \textbf{document}?                &
What type of product is on the \textbf{image}?               &
Which company \textbf{appears} first among the attendees?    &
What is the \textbf{title} of the diagram?                   \\
  &  &  &  \\
  
What ‘council’ is mentioned in the \textbf{document}?           &
In the \textbf{image} of the playing card pack, what is the number on the card of diamonds? &
Which is the numerical rating that \textbf{appears} most number of times? &
Who prepared the controversial report en\textbf{title}d "Dietary Goals for the United States"? \\
  &  &  &  \\
  
Which date is mentioned at the end of the ‘\textbf{document}’? &
What is the name of the company in the \textbf{image}?       &
Which is the page number greater than 28, that \textbf{appears} only once?   &
What is the \textbf{title} of this page?                     \\
\bottomrule

\end{tabular}}
\vspace{1mm}
\caption{Key-words used to find inadequate questions over multi-page documents. In the title row, following each key-word is showed the number of questions in SingleDocVQA with that word.}
\label{tab:key-word_analysis}
\end{table*}





\newpage

\section{Document pages during training} \label{appendix:train_doc_pages}

As described in \cref{sec:method}, it is not feasible to train with 20 page length documents due to training resource limitations. However, as we show in \cref{tab:train_pages}, even though the model performs significantly worse when trained with a single page, the returns become diminishing when training with more than 2. Thus, as explained in \cref{sec:method} we decided to use 2 pages in the first stage of training.

\begin{table}[H]
\centering
\small
\begin{tabular}{ccc}
\toprule
Trained pages & Acc   & ANLS   \\
\midrule
1             & 22.96 & 0.3860 \\
2             & 33.37 & 0.5577 \\
5             & 34.08 & 0.5730 \\
10            & 34.25 & 0.5792 \\
\bottomrule
\end{tabular}
\caption{Experiments showing the results when training with different number of document pages and tested with the document original length.}
\label{tab:train_pages}
\end{table}

\newpage
\section{Hyperparameters} \label{appendix:hyperparameters}

\begin{table}[H]
\resizebox{\columnwidth}{!}{
\begin{tabular}{lrrrrr} \toprule
                      & \multicolumn{1}{c}{\textbf{BERT}} &
                      \multicolumn{1}{c}{\textbf{Longformer}} &
                      \multicolumn{1}{c}{\textbf{BigBird}} &
                      \multicolumn{1}{c}{\textbf{T5}} &
                      \multicolumn{1}{c}{\textbf{\methodName}\textsuperscript{$\dagger$}} \\
\midrule
Model size            & large       & base      & base      & base      & base      \\
Parameters            & 334M        & 148M      & 131M      & 223M      & 316M      \\
Model initial weigths & 
\href{https://github.com/mineshmathew/DocVQA/tree/master/BERT_baseline}{SingleDocVQA} & \href{https://huggingface.co/valhalla/longformer-base-4096-finetuned-squadv1}{SQuADv1}  & \href{https://huggingface.co/google/bigbird-base-trivia-itc}{TrivaQA}   & 
\href{https://huggingface.co/t5-base}{C4} &
\href{https://huggingface.co/t5-base}{C4}                               \\
Max Seq. Length       & 512         & 4096      & 4096      & 512       & 20480     \\
Training Loss         & CE          & CE        & CE        & CE        & CE        \\
batch size            & 32          & 8         & 8         & 20        & 8         \\
lr                    & 5e-5        & 1e-4      & 3e-5      & 2e-4      & 2e-4      \\
optimizer             & AdamW       & AdamW     & AdamW     & AdamW     & AdamW     \\
scheduler             & linear      & linear    & linear    & linear    & linear    \\
warmup iterations     & 1000        & 1000      & 1000      & 1000      & 1000      \\
training epochs       & 1           & 10        & 10        & 10        & 1 - 10 - 1    \\
\bottomrule
\end{tabular}}
\caption{Hyperparameters of the baselines and the proposed method that were used to train and evaluate on \datasetName. $\dagger$:~ \methodName refers to all three pre-training, training and fine-tune stages. The only difference is the number of epochs: 1, 10 and 1 respectively. Training loss CE denotes CrossEntropy loss.}
\label{tab:training_parameters}
\end{table}

\begin{figure*}[ht]
  \includegraphics[width=\textwidth]{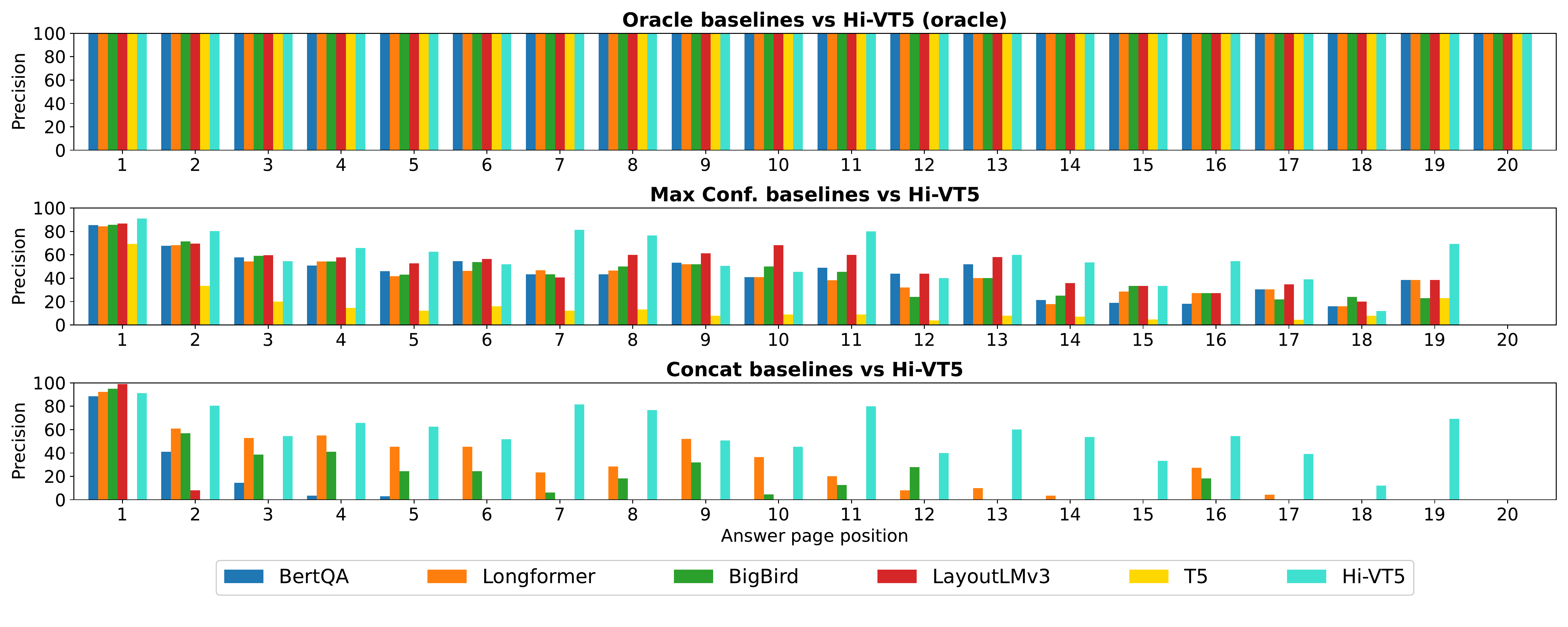}
  \caption{\textbf{Accuracy of page identification as a function of answer page position}. The figure shows the page identification accuracy of the different baselines and \methodName in the oracle \setup (top), and the baselines in the \logits (middle) and concat (bottom) \setup against \methodName using the page identification module. Notice that the breakdown of the scores is NOT performed on the number of pages the document, but in which page the answer is found.}
  \label{fig:methods_ret_prec_by_answer_page}
\end{figure*}


\newpage

\section{Page identification accuracy by answer page position}

In \cref{fig:methods_ret_prec_by_answer_page} we show the answer page identification accuracy of the different baselines and the proposed method, as a function of the page number of the answer. The overall performance follows a similar behavior as the answer scores. \longformer is the baseline that performs the best in the concat setting, and and the performance gap between this and the rest of the baselines becomes more significant as the answer page is located in the final pages of the document. However, \methodName outperforms all the baselines by a big margin.

\newpage
\section{\methodName attention visualization} \label{appendix:attention_viz}

\begin{figure*}[htbp]
    \centering
    \hfill
    \begin{subfigure}[b]{0.42\textwidth}
        \centering
        \includegraphics[width=\textwidth]{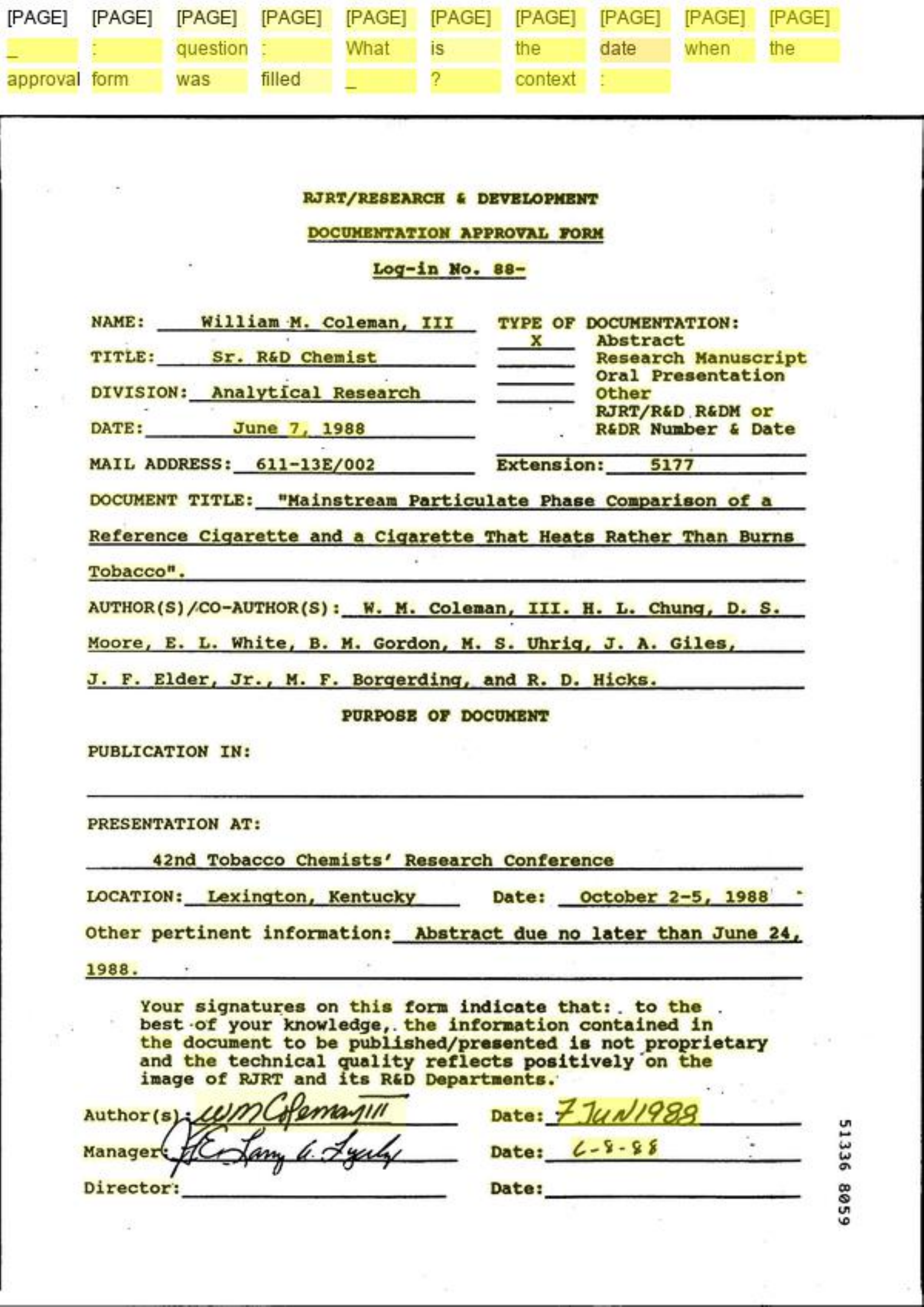}
        \caption{Global attention over all the text in the page}
        \label{subfig:att_global}
    \end{subfigure}
    \hfill
    \begin{subfigure}[b]{0.42\textwidth}  
        \centering 
        \includegraphics[width=\textwidth]{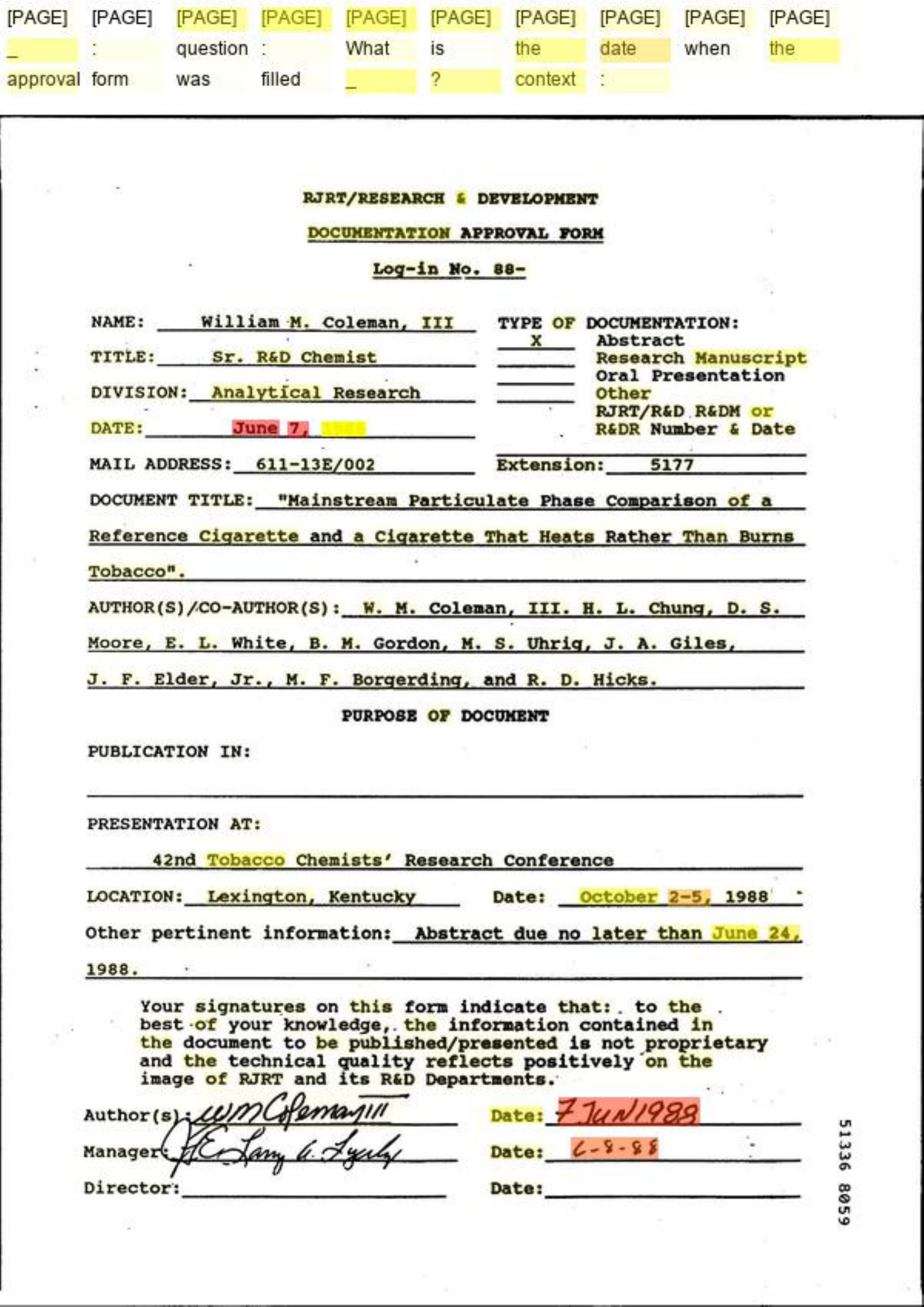}
        \caption{Attention focused over the OCR tokens corresponding to the answer (7 June, 1988)}
        \label{subfig:att_answer1}
    \end{subfigure} 
    \hfill  \\
    
    \hfill
    \begin{subfigure}[b]{0.42\textwidth}   
        \centering 
        \includegraphics[width=\textwidth]{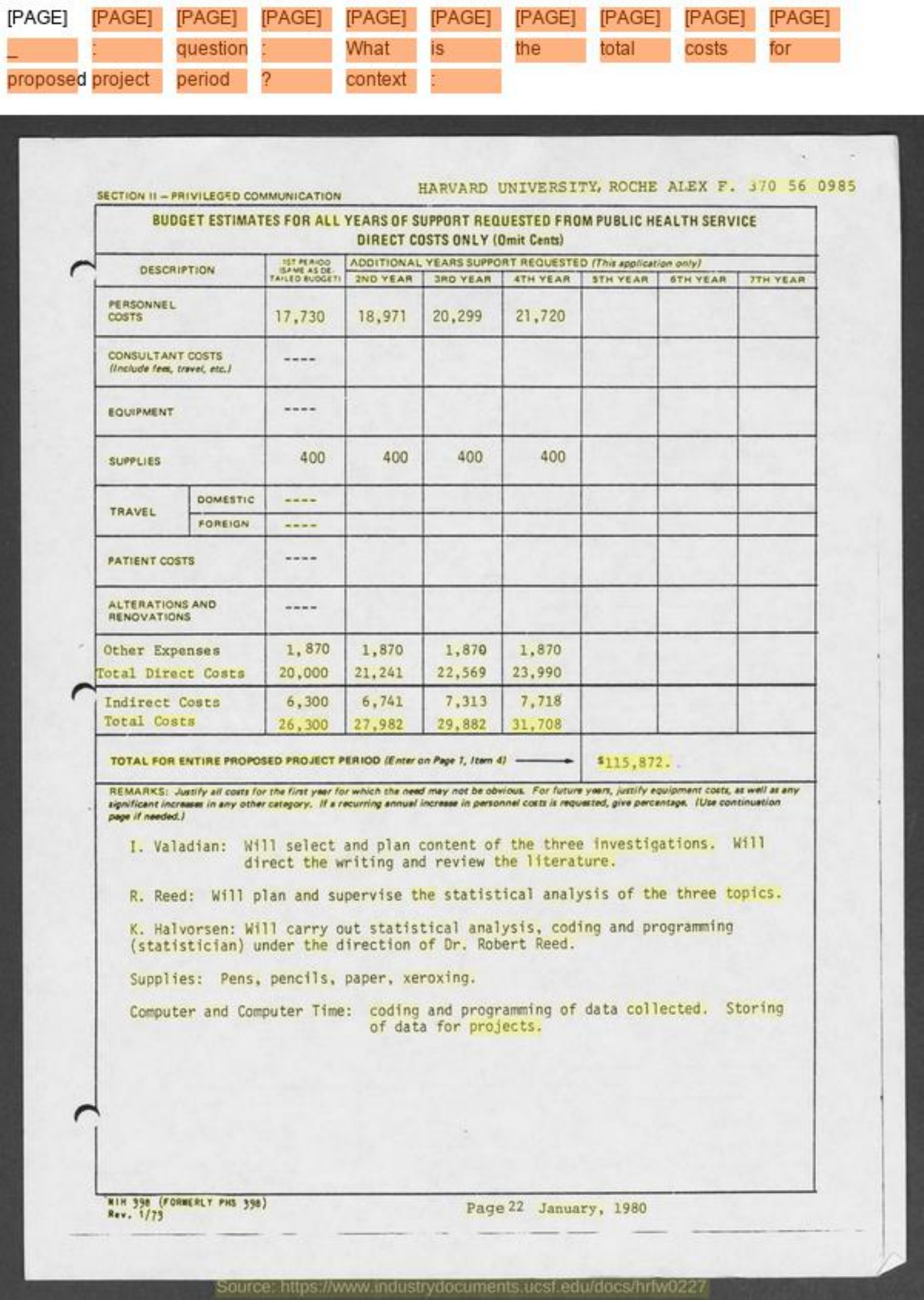}
        \caption{Attention focused over the rest of the \pageToken and question tokens.}
        \label{subfig:att_question}
    \end{subfigure}
    \hfill
    \begin{subfigure}[b]{0.42\textwidth}   
        \centering 
        \includegraphics[width=\textwidth]{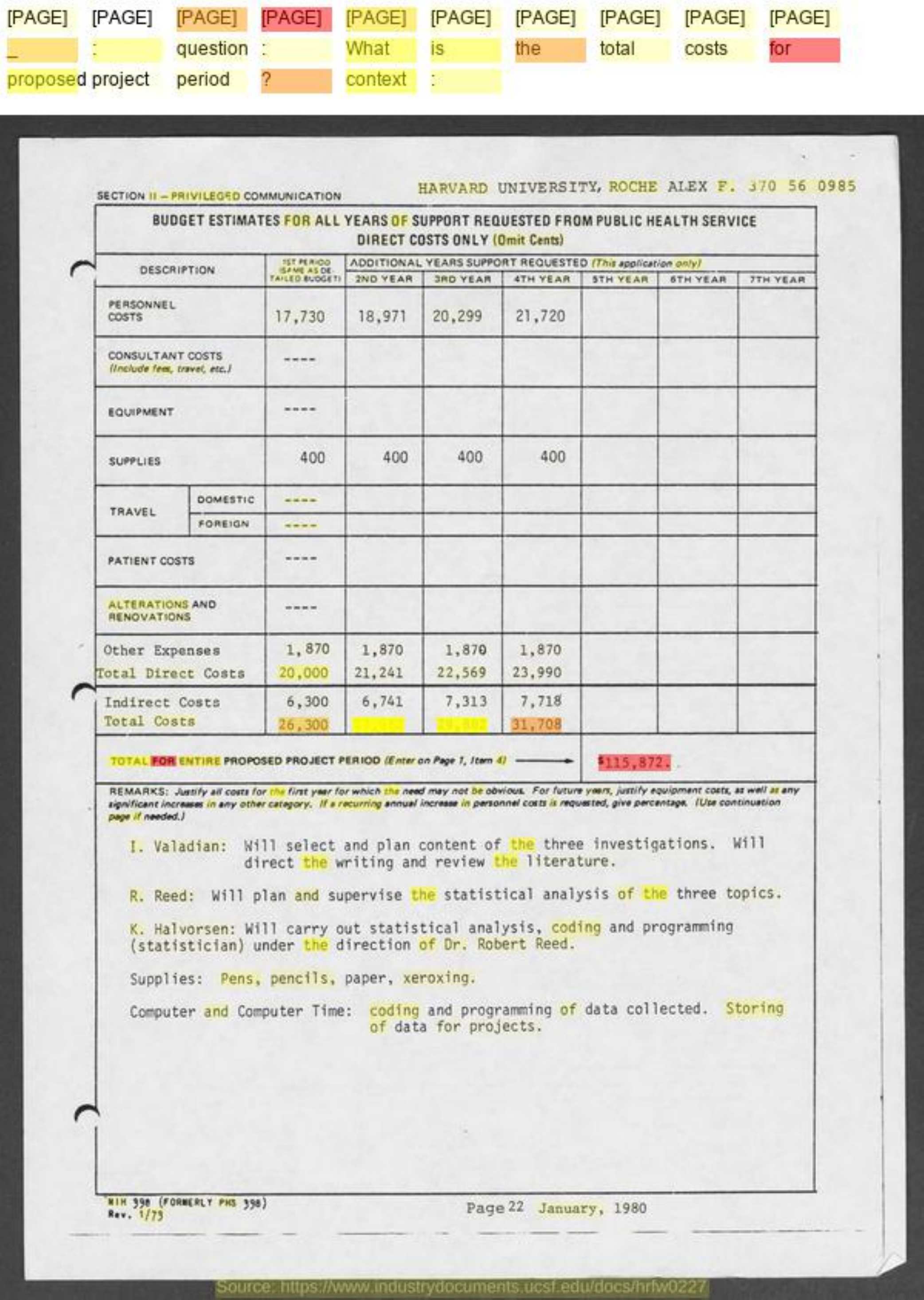}
        \caption{Attention focused over the OCR tokens corresponding to the answer (\$115.872)}
        \label{subfig:att_answer2}
    \end{subfigure}
    \hfill
    \caption{Visualization of the \methodName attention scores.}
    \label{fig:att_viz}

\end{figure*}

To further explore the information that \methodName embeds into the \pageToken tokens, we show the attention scores 
for some examples in \datasetName. The attention of \cref{subfig:att_global}, corresponds to the first \pageToken token, which usually performs a global attention over the whole document with a slight emphasis on the question tokens, which provides a holistic representation of the page. Other tokens like in \cref{subfig:att_question} focuses its attention over the other \pageToken, and question tokens. More importantly, there is always a token that focuses its attention to the provided answer like in Figs. \ref{subfig:att_answer1} and \ref{subfig:att_answer2}.
 
\end{document}